\documentclass{article}

\usepackage{arxiv}

\usepackage[utf8]{inputenc} 
\usepackage[T1]{fontenc}    

\usepackage{url}            
\usepackage{booktabs}       
\usepackage{amsfonts}       
\usepackage{nicefrac}       
\usepackage{microtype}      
\usepackage{lipsum}
\usepackage{graphicx}
\usepackage{xurl}                 
\usepackage[hidelinks]{hyperref}  
\usepackage{etoolbox}
\apptocmd{\thebibliography}{\raggedright\setlength{\emergencystretch}{3em}}{}{}

\usepackage{balance}

\usepackage{cite}
\usepackage{amsmath,amssymb,amsfonts}
\usepackage{algorithmic}
\usepackage{graphicx}
\usepackage{textcomp}
\usepackage{xcolor}
\usepackage{subfigure}
\def\BibTeX{{\rm B\kern-.05em{\sc i\kern-.025em b}\kern-.08em
    T\kern-.1667em\lower.7ex\hbox{E}\kern-.125emX}}
\graphicspath{ {./images/} }

\title{An Explainable and Fair AI Tool for PCOS Risk Assessment: Calibration, Subgroup Equity, and Interactive Clinical Deployment}

\author{
 Asma Sadia Khan \\
  Department of Biomedical Engineering\\
  Chittagong University of Engineering and Technology\\
  Chittagong, Bangladesh \\
  \texttt{u1911007@student.cuet.ac.bd} \\
   \And
 Sadia Tabassum \\
  Department of Biomedical Engineering\\
  Chittagong University of Engineering and Technology\\
  Chittagong, Bangladesh \\
  \texttt{u1911029@student.cuet.ac.bd} \\
   \
}

\begin{document}
\maketitle
\begin{abstract}
This paper presents a fairness-audited and interpretable machine learning framework for predicting polycystic ovary syndrome (PCOS), designed to evaluate model performance and identify diagnostic disparities across patient subgroups. The framework integrated SHAP-based feature attributions with demographic audits to connect predictive explanations with observed disparities for actionable insights. Probabilistic calibration metrics (Brier Score and Expected Calibration Error) are incorporated to ensure reliable risk predictions across subgroups. Random Forest, SVM, and XGBoost models were trained with isotonic and Platt scaling for calibration and fairness comparison. A calibrated Random Forest achieved a high predictive accuracy of 90.8\%. SHAP analysis identified follicle count, weight gain, and menstrual irregularity as the most influential features, which are consistent with the Rotterdam diagnostic criteria. Although the SVM with isotonic calibration achieved the lowest calibration error (ECE = 0.0541), the Random Forest model provided a better balance between calibration and interpretability (Brier = 0.0678, ECE = 0.0666). Therefore, it was selected for detailed fairness and SHAP analyses. Subgroup analysis revealed that the model performed best among women aged 25–35 (accuracy 90.9\%) but underperformed in those under 25 (69.2\%), highlighting age-related disparities. The model achieved perfect precision in obese women and maintained high recall in lean PCOS cases, demonstrating robustness across phenotypes. Finally, a Streamlit-based web interface enables real-time PCOS risk assessment, Rotterdam criteria evaluation, and interactive ‘what-if’ analysis, bridging the gap between AI research and clinical usability.
\end{abstract}


\section{\textbf{Introduction}}

Polycystic Ovary Syndrome (PCOS) is one of the most prevalent endocrine disorders, affecting approximately 6–13\% of women of reproductive age, yet up to 70\% of those affected remain undiagnosed worldwide \cite{noauthor_polycystic_nodate}.
Characterized by irregular menstrual cycles, metabolic irregularities, Hyperandrogenism, and polycystic ovarian morphology, PCOS not only affects fertility but is also linked to long-term risks such as type 2 diabetes, cardiovascular issues, obesity, osteoporosis, endometrial cancer, depression, etc \cite{upreti2025polycystic}\cite{rasquin2017polycystic}. 
Recent advances in machine learning (ML) have shown promise in supporting clinical decision-making for PCOS diagnosis, particularly when routine demographic and laboratory data are used \cite{barrera2023application}. However, many existing models are developed in a black-box fashion, prioritizing predictive accuracy without evaluating the reliability and fairness of the predicted probabilities across different patient subgroups. This is particularly critical in women's health, where systemic biases in data collection and care delivery can lead to disparities in diagnosis and treatment \cite{cherian2024statistical}.

Moreover, while model performance metrics such as accuracy and AUC offer general insights, they do not reflect how well a model's predicted probabilities align with real-world outcomes, which is a key consideration for actionable clinical use.
In this study, we present an Uncertainty-Aware and Fair Machine Learning Framework for PCOS prediction using publicly available clinical data. We train and calibrate three popular models: Random Forest, Support Vector Machine (SVM), and XGBoost, using both Platt scaling and Isotonic regression. Beyond overall performance, we analyze subgroup fairness (based on factors such as age, BMI, and pregnancy status), feature importance using SHAP, and model calibration through the Brier Score and Expected Calibration Error (ECE). 
Our study demonstrates that a calibrated Random Forest model achieves high predictive accuracy for PCOS (up to 90.8\%) while offering robust clinical interpretability. SHAP analysis revealed that follicle count, weight gain, and menstrual irregularity were the most influential features, aligning with Rotterdam diagnostic criteria \cite{eshre2004revised}. While SVM with isotonic calibration achieved the best calibration (ECE: 0.0541), Random Forest offered the best trade-off between calibration and explainability (Brier: 0.0678, ECE: 0.0666), and was therefore selected for fairness and SHAP-based audits. Subgroup analysis revealed that the model performed best in women aged 25–35 (Accuracy: 90.9\%) but underperformed in women under 25 (Accuracy: 69.2\%), indicating age-related diagnostic disparities. Obese women had perfect precision, while lean PCOS cases were still detectable with high recall, supporting the model’s robustness across phenotypes.

We developed a web-based prototype that enhances conventional PCOS risk calculators by integrating fairness auditing, explainability, and clinical guideline alignment.

\begin{itemize}
        \item Developed a fairness-audited, interpretable PCOS prediction framework that evaluates model performance across relevant subgroups to identify diagnostic inequities.
        \item Linked feature attribution to subgroup disparities by combining SHAP-based explanations with demographic audits, enabling actionable insights.
        \item Integrated probabilistic calibration using Brier Score and Expected Calibration Error (ECE) to ensure clinically reliable predicted probabilities across subgroups.
        \item Benchmarked multiple models and calibration methods (Random Forest, SVM, XGBoost with isotonic and Platt scaling) to compare accuracy, calibration, and fairness.
        \item Designed a user-friendly, web-based interface using Streamlit, providing real-time risk assessment, Rotterdam criteria evaluation, and "What-If" analysis, making health insights accessible without local installation.
    \end{itemize}

\section{\textbf{Literature Review}}
Recent advancements in machine learning, deep learning, and artificial intelligence have significantly enhanced the accuracy and efficiency of PCOS diagnosis.

Elmannai et al.\cite{elmannai2023polycystic} developed a stacking ensemble model for early PCOS detection, combining LR,  DT, RF,  SVM, KNN, NB, XGBoost, and AdaBoost as base learners with RF as the meta-learner. SMOTEENN handled class imbalance, RFE/tree-based/mutual information methods selected features, and Bayesian optimization tuned hyperparameters. The stacking model with RFE feature selection achieved the highest performance, with an accuracy of 98.87\%. In a study, Wang et al.\cite{wang2025artificial} provided a comprehensive systematic review of AI applications in PCOS management, showing performance comparable to or better than that of clinicians in PCOS diagnosis and prediction. By integrating multi-omics bioinformatics, clinical, and imaging data, AI enables patient stratification, personalized care, and improved screening. Although still in early clinical adoption, AI is expected to enhance efficiency, interpretability, and accessibility, especially in resource-limited settings. However, the review lacks discussion on model interpretability, reliability, and subgroup fairness, which are vital for safe and equitable clinical use. 

Jain et al.\cite{jain2024xplainable} proposed a PCOS prediction model using GNB, kNN, SVM, RF, and DNN, achieving 97\% accuracy. Explainable AI techniques (LIME, SHAP) identified insulin level, follicle no (R), and follicle no (L) as key factors affecting PCOS, enabling an interpretable and trustworthy diagnostic approach. Mahesswari et al. \cite{mahesswari2024smartscanpcos} introduced an XAI-based PCOS predictor structured as a hierarchical two-tier Random Forest ensemble, developed after evaluating seven conventional classifiers and two stacking ensembles. Using an open-source dataset (Kaggle dataset) with features selected via TOPCA, OSSM, and TOMIM, it achieved 99.31\% accuracy with 17 TOMIM-selected features and 99.32\% with 8-fold cross-validation. Shapash was used to provide transparent and interpretable predictions through XAI visualizations.

Divenkar et al.\cite{divekar2024leveraging} developed a deep learning approach using the InceptionV3 architecture for ultrasound-based PCOS diagnosis. The model achieved 90.52\% accuracy, 97.16\% recall, 90.01\% precision, and a 93.45\% F1-score, with LIME and saliency map enhancing interpretability. Despite dataset size limitations, the study demonstrated the potential of transfer learning to develop robust and clinically relevant diagnostic tools. Chelliah et al. \cite{chelliah2024enhancing} applied nine machine learning models (ELM, Isolation Forest, FM, MGP, NMF, GP, DBN, PSO, LSTM) with Explainable AI (SHAP, LIME, Qlattice) to improve PCOS detection. Their results show that combining diverse ML methods can enhance early diagnosis and support personalized treatment. \cite{moral2024podboost} developed a four-phase ML framework for early PCOS prediction using demographic and clinical data. SMTL was applied for data balancing, GWO for feature selection, and a hybrid PODBoost classifier for prediction, with LIME used for interpretability. The model achieved 97.42\% accuracy, showing strong potential for PCOS and other multimodal disease prediction.

Recent PCOS prediction studies show ML models can exceed 97\% accuracy using ensemble, deep learning, or hybrid metaheuristics like Gray Wolf Optimization (GWO) and PODBoost. However, many lack evaluation of probabilistic calibration, subgroup fairness, and clinical interpretability. They often omit metrics like Brier score, Expected Calibration error, or subgroup analysis by age, BMI, or lifestyle factors. Additionally, few studies align their explainability outputs with established diagnostic guidelines such as the Rotterdam criteria.
Our work presents a probability-calibrated and demographically audited PCOS prediction framework, using SHAP explanations grounded in clinical rationale. It achieves high accuracy ($~$91\%) with reliable probability estimates (ECE = 0.0541) and conducts subgroup fairness audits, identifying performance gaps in younger and leaner PCOS phenotypes. By focusing on transparency, reproducibility, and equity, our approach addresses key gaps in existing research and enhances the practical readiness of AI-assisted PCOS screening tools.

\section{\textbf{Methodology}}
\subsection{Dataset Description}
We used a publicly available dataset that includes comprehensive physical and clinical parameters related to PCOS and infertility. The data were collected from 10 different hospitals across Kerala, India \cite{Kaggle_PCOS_2022}. It consists of 541 instances and 40 features, with clinical, anthropometric, hormonal, lifestyle, and ultrasound-based measurements. To evaluate the model’s fairness and subgroup generalization, 
\subsection{Data Pre-processing}
The target variable, PCOS (Y/N), indicating the presence (1) or absence (0) of PCOS, was encoded as a binary integer using type conversion. The features (X) were separated from the target (y) to prepare for further processing and modeling. Rows containing missing values were removed to ensure a complete dataset. Feature selection was performed by excluding the target variable and any identifier columns. The remaining features were divided into two subsets:
\begin{itemize}
    \item \textbf{Binary (Categorical) Features:} The identified binary columns were kept as unscaled categorical variables (0/1).
    \item \textbf{Continuous Features:} All other numeric columns ( Age (yrs), BMI, FSH(mIU/mL)) were selected for scaling.
\end{itemize}

\subsection{Train-Test Split}
The feature matrix (X) and target (y) were split into training (80\%) and test (20\%) sets using stratified sampling to maintain the class distribution of PCOS (Y/N).

\subsection{Selective Feature Scaling}
For preserving the categorical nature of binary features, only the continuous features were standardized after the train–test split \cite{article}. The StandardScaler from scikit-learn was fitted on the continuous features of the training set to calculate their mean and standard deviation, ensuring no information from the test set was used. 

\subsection{ Model Training}
Three classifiers were trained:
\begin{itemize}
    \item Random Forest (RF) using 5-fold cross-validation with grid search over depth and number of estimators.
    \item Support Vector Machine (SVM) with RBF kernel, and XGBoost (XGB) classifier with tuned max depth and estimators.
    \item  Model training was performed using the scikit-learn and XGBoost libraries with fixed random seeds for reproducibility.
\end{itemize}

\subsection{Probabilistic Calibration}
To improve the reliability of predicted probabilities, two calibration methods were applied to Random Forest, SVM, and XGBoost classifiers using the CalibratedClassifierCV class from scikit-learn with 5-fold cross-validation. The methods used were Platt Scaling and Isotonic Regression \cite{niculescu2005predicting}, applied independently to each classifier. 

\begin{itemize}
    \item \textbf{Isotonic Regression (Non-Parametric Calibration):}  Isotonic regression can be applied in probabilistic classification to calibrate the predicted probabilities of supervised machine learning models.
    It allows for flexible adjustments to the probability distributions, effectively handling complex, non-linear distortions, particularly in datasets with non-uniform distributions, such as medical data \cite{jiang2011smooth}.
    
    \item \textbf{Platt Scaling (Sigmoid-Based Parametric Calibration):} Platt scaling, also known as Platt calibration, is a technique used to transform the outputs of a classification model into a probability distribution over classes. 
    It was selected for its simplicity and effectiveness in binary classification tasks \cite{platt1999probabilistic}.
\end{itemize}
Both calibration methods were evaluated independently to compare their impact on probability reliability.

\section{\textbf{Evaluation Metrics}}
To check how good our PCOS prediction models are and how useful they are in clinical settings, we used several ways to measure their performance: Accuracy, Brier Score, Expected Calibration Error (ECE), Calibration Curves, and Decision Curve Analysis (DCA). These measures help us understand how well the model separates cases (discrimination), how reliable its predictions are (probabilistic reliability), how well predicted risks match actual outcomes (calibration), and how helpful it is for medical decisions (clinical utility).







\subsection{Brier Score:}
Measures the accuracy of predicted probabilities, with lower values indicating better calibration \cite{liu2022deep}.
\[
\mathrm{Brier}=\frac{1}{N}\sum_{i=1}^{N}(p_i - y_i)^2,\quad \mathrm{Brier}\in[0,1],
\]

\subsection{Expected Calibration Error (ECE):}
Expected Calibration Error (ECE) evaluates how well a model’s estimated probabilities reflect the observed probabilities by taking a weighted average over the absolute difference between average accuracy (acc) and average confidence (conf) \cite{pavlovic2025understanding}. It involves splitting all \( n \) datapoints into \( M \) equally spaced bins:

\[
\text{ECE} = \sum_{m=1}^{M} \frac{|B_m|}{n} \left| \text{acc}(B_m) - \text{conf}(B_m) \right|,
\]

where \( B \) is used to represent the "bins" and \( m \) for the bin number, while \( \text{acc} \) and \( \text{conf} \) are:

\[
\text{acc}(B_m) = \frac{1}{|B_m|} \sum_{i \in B_m} 1(\hat{y}_i = y_i)
\]

\quad \text{and} \quad

\[
\text{conf}(B_m) = \frac{1}{|B_m|} \sum_{i \in B_m} \hat{p}(x_i)
\]

\noindent
where,

\( \hat{y}_i \) is the model’s predicted class (\emph{arg max}) for sample \( i \)

\( y_i \) is the true label for sample \( i \). 

1 is an indicator function, meaning when the predicted label \( \hat{y}_i \) equals the true label \( y_i \), it evaluates to 1; otherwise 0.

\subsection{Calibration Curves:}
Calibration plot the actuals probability against the predicted probability \cite{noauthor_calibration_nodate}. A well-calibrated model will have plotted points that lie closer to the diagonal line. A calibration slope << 1 is indicative of overfitting a model to the test data.

\subsection{Decision Curve Analysis (DCA):}
Decision Curve Analysis (DCA) is a method used to evaluate the clinical value of prognostic or diagnostic models, decision rules, or biomarkers \cite{piovani2023optimizing}. Clinical utility is measured using net benefit, which is calculated using the following formula: 
\[
\text{Net benefit} = \left(\frac{\text{TP}}{n}\right) - \left(\frac{\text{FP}}{n}\right) \times \left( \frac{P_t}{1 - P_t} \right)
\]
Where:

 \({P_t} \) : probability threshold at which a clinician decides to take action based on the predicted outcome.

n : total number of patients.

TP, FP : number of true and false positives at threshold \({P_t} \).

 \section{\textbf{Fairness Auditing Methodology}}

To ensure equitable model performance across clinically and demographically relevant subpopulations, we conducted a structured fairness audit using stratified evaluation across multiple sensitive features. This section outlines the preprocessing, subgroup definition, and evaluation strategy.
We used the Random Forest model for subgroup fairness and SHAP explainability analyses due to its strong interpretability, native support for SHAP decomposition, and competitive overall calibration. 

\subsubsection{Sensitive Feature Extraction and Grouping:}
Below, we outline the features selected for subgroup analysis, how they were grouped, and their clinical significance:

\begin{table*}[h!]
\centering
\caption{Stratification of groups, ranges, rationale, and clinical representation for PCOS analysis.}
\renewcommand{\arraystretch}{1.3}
\begin{tabular}{|p{2cm}|p{3cm}|p{6cm}|p{5cm}|}
\hline
\textbf{Group} & \textbf{Range} & \textbf{Rationale} & \textbf{Clinical Representation} \\ \hline

Age &
$<25$, 25--35, $>35$ &
Symptom onset/diagnosis differ by age: 25--35 = peak reproductive/diagnosis years; $<25$ often underdiagnosed; $>35$ influenced by aging endocrine changes \cite{Roe2011Diagnosis}. &
$<25$: Adolescent/early adult. \newline 25--35: Peak diagnosis. \newline $>35$: Late reproductive/perimenopause. \\ \hline

BMI &
Normal $<25$, Overweight 25--29.9, Obese $\geq 30$ &
Obesity/insulin resistance affect severity and phenotype; stratification shows metabolic burden effects \cite{amisi2022markers}. &
Normal: Lean phenotype. \newline Overweight: Moderate risk. \newline Obese: High-risk, insulin resistant. \\ \hline

Pregnancy &
Pregnant / Not Pregnant &
Pregnancy hormones may obscure PCOS signals; comparison tests robustness across hormonal baselines \cite{Palomba2015Pregnancy}. &
Pregnant: Hormonal confounding. \newline Not Pregnant: Standard baseline. \\ \hline

Marital Duration &
$\leq 5$ yrs, $>5$ yrs &
Proxy for reproductive stage and likelihood of fertility workup. &
$\leq 5$: Early reproductive phase. \newline $>5$: Possible infertility history. \\ \hline

Lifestyle &
Exercise (Y/N), Fast Food (Y/N) &
Adjustable risk factors; fairness across lifestyle subgroups. &
Exercise: Insulin sensitivity. \newline Fast Food: Poor diet/metabolic burden. \\ \hline

\end{tabular}
\end{table*}

\subsection{Data Preprocessing and Group Alignment}
To ensure clean subgroup comparisons, unnecessary or irrelevant columns, such as ID columns, were removed. Samples with missing values were excluded to maintain data integrity. Numeric features were standardized using z-score normalization to facilitate fair model training. Target labels were encoded as binary ( 0 = No PCOS, 1 = PCOS). Separate metadata splits were created for subgroup evaluation on the test set to prevent information leakage.

\subsection{Fairness Evaluation via Subgroup Metrics}
A Random Forest classifier was trained on the normalized training data and evaluated on the holdout test set. We used the fairlearn library’s MetricFrame \cite{weerts2023fairlearn}. To assess performance on the following key metrics, disaggregated by each sensitive feature: Accuracy for overall correctness within the subgroup, Precision, and Recall.

This approach allowed us to assess differences in model performance across age, BMI, lifestyle, and reproductive history, revealing areas where the model performs well and where it may under-represent certain clinical phenotypes.

\subsection{SHAP Analysis}
To interpret the model’s predictions, SHAP values were computed using the TreeExplainer from the SHAP library, tailored for tree-based models like Random Forest \cite{hasan2023understanding}. SHAP values were calculated for the test set, focusing on the positive class (PCOS, class 1) for binary classification. A summary plot was generated to visualize global feature importance.

\subsection{Prototype Implementation for Clinical and Patient Use}
We developed an interactive prototype using the Streamlit framework. 

The core features are:

\begin{itemize}
    \item \textbf{Risk Prediction Panel:} Displays the predicted PCOS probability, calibrated using Isotonic or Platt methods.
    \item \textbf{Fairness \& Confidence Flags:} Highlights model performance in relevant subgroups (age, BMI, pregnancy status), ensuring transparency.
    \item \textbf{SHAP-based Explainability:}
    
\textbf{Clinician View:} Ranked feature contributions with raw values and clinical context.

\textbf{Patient View:} Simple explanation of the top three contributing factors.

    \item \textbf{Rotterdam Criteria Alignment:} Automatically evaluates whether the profile meets $\geq$2 of the 3 Rotterdam criteria (oligo/anovulation, hyperandrogenism, polycystic ovarian morphology) \cite{practitioners_polycystic_nodate} using available dataset features, and displays a clear statement on whether the requirements are met.

    \item \textbf{Supportive Indicators:} Displays relevant non-diagnostic indicators (AMH, BMI, TSH) to contextualize risk.

    \item \textbf{What-If Analysis:} Allows adjustments of features (weight, exercise, cycle regularity) to see potential impacts on risk, aiding patient counseling and decision making.
\end{itemize}




\section{\textbf{Results}}
\subsection{Model Performance Comparison}
To assess the reliability and clinical usefulness of predicted probabilities for PCOS diagnosis, we evaluated each model using three key metrics: Accuracy, Brier score, and ECE given in Table \ref{tab:1}.

\begin{table}[ht]
    \centering
    \caption{Model Performance Comparison}
    \begin{tabular}{|c|c|c|c|c|}
    \hline
        \textbf{Model} & \textbf{Calibration} & \textbf{Accuracy} & \textbf{Brier Score} & \textbf{ECE} \\
        \hline
        SVM & Isotonic & 0.8991 & 0.0764 & 0.0541 \\
        Random Forest & Isotonic & 0.8991 & 0.0678 & 0.0666 \\
        XGBoost & Isotonic & 0.8899 & 0.0733 & 0.0663 \\
        SVM & Platt & 0.8899 & 0.0831 & 0.0779 \\
        Random Forest & Platt & 0.9083 & 0.0713 & 0.0813 \\
        XGBoost & Platt & 0.9083 & 0.0717 & 0.0957 \\
        \hline
    \end{tabular}
    \label{tab:1}
\end{table}

\subsection{Interpretation of Figures}

\subsubsection{Calibration Curves}

\begin{figure}[htbp]
  \centering
  \includegraphics[width=0.8\linewidth]{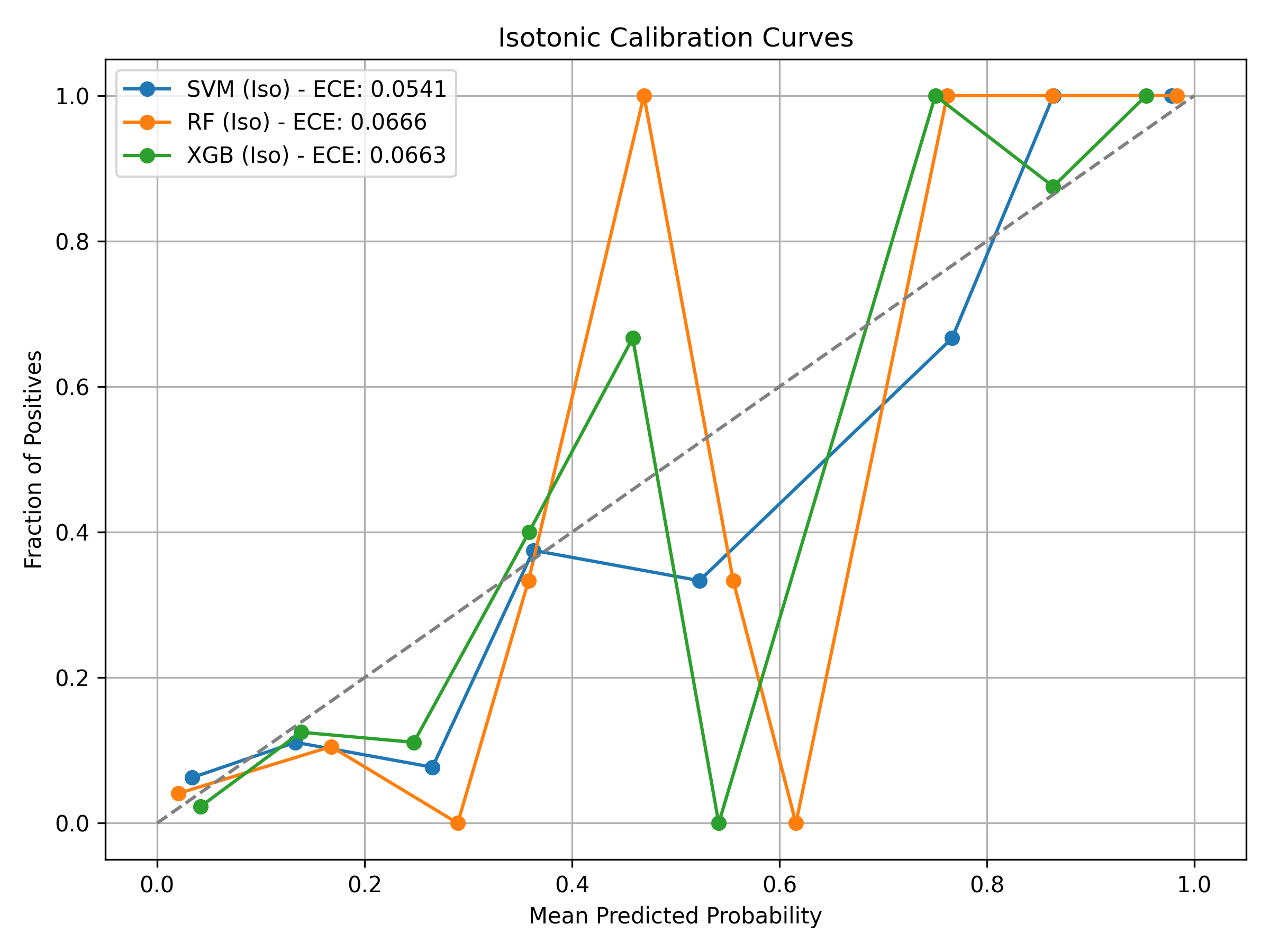}
  \label{fig:1}
  \caption{Isotonic Claibration Plot (SVM, RF, XGB)}
\end{figure}

Isotonic calibration (Fig. \ref{fig:1}) showed better overall alignment with the ideal diagonal calibration line, particularly for SVM (Iso), which achieved the lowest ECE.

\begin{figure}[htbp]
  \centering
  \includegraphics[width=0.8\linewidth]{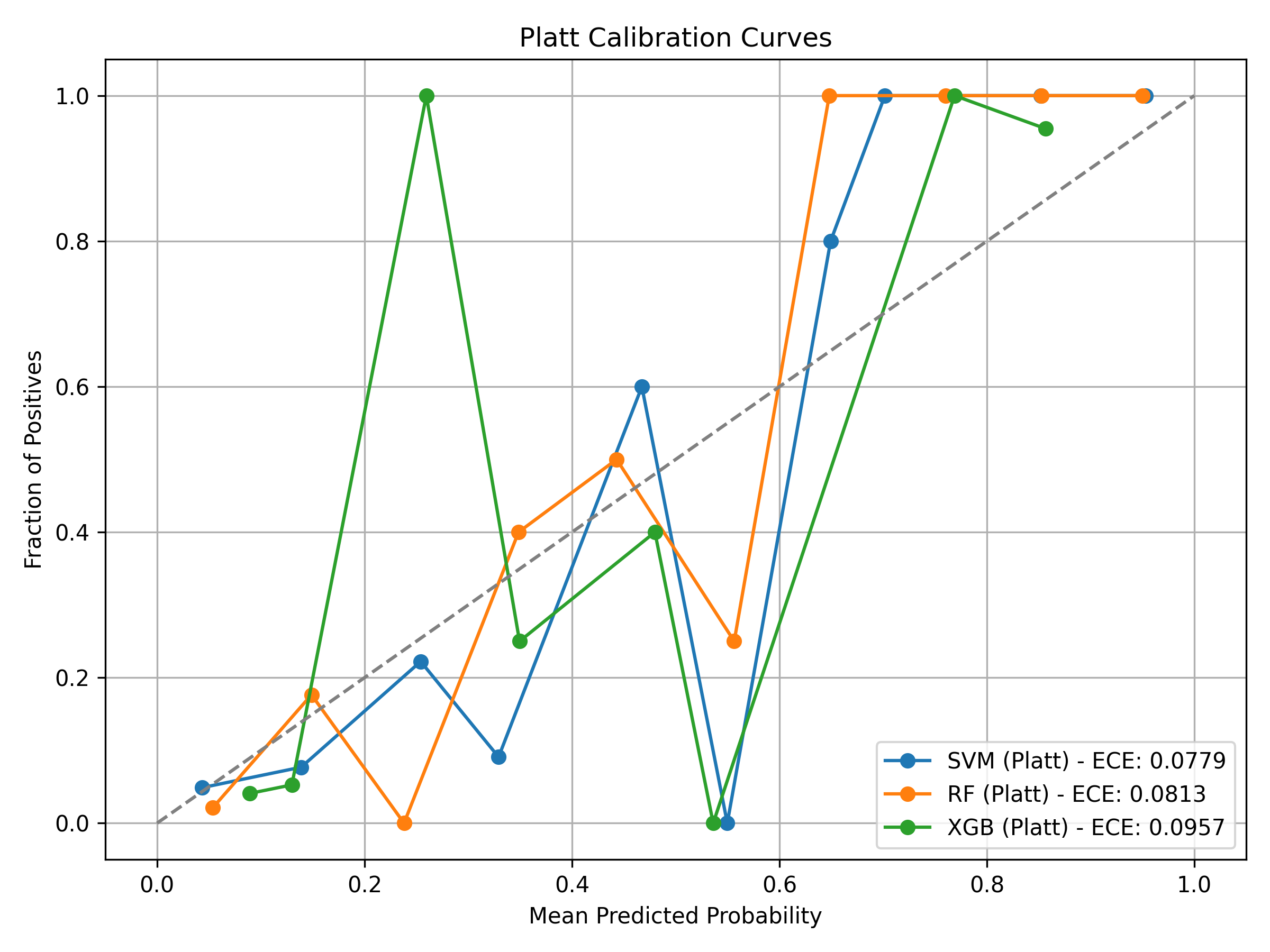}
  \label{fig: 2}
  \caption{Platt Calibration Plot (SVM, RF, XGB)}
\end{figure}

Platt calibration (Fig. \ref{fig: 2}) introduced greater variability, especially in XGBoost, which exhibited the largest deviation from perfect calibration and the highest ECE (0.0957).

\subsubsection{Decision Curve Analysis}

\begin{figure}[htbp]
  \centering
  \includegraphics[width=0.8\linewidth]{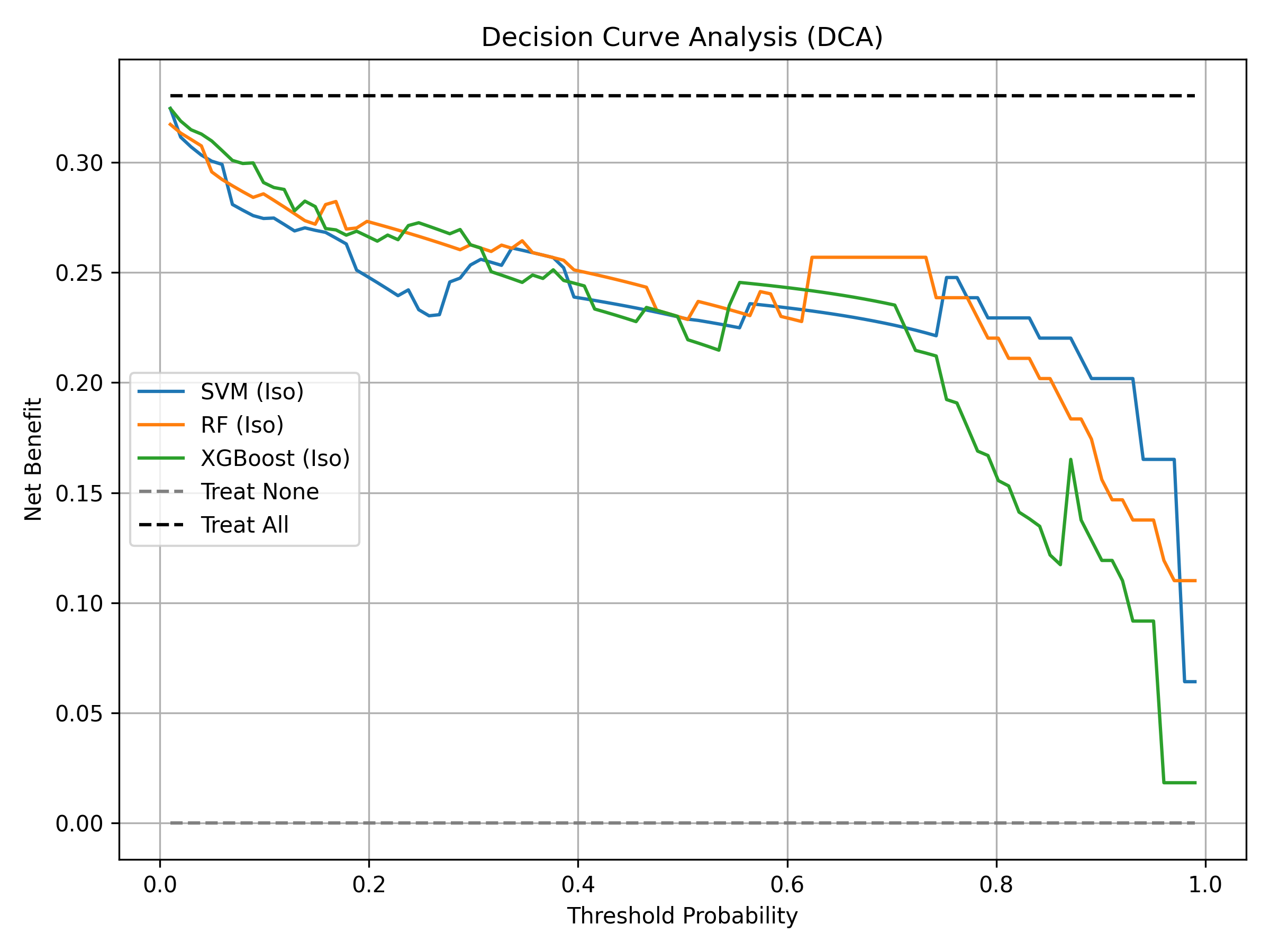}
  \label{fig:3}
  \caption{Decision Curve Analysis (SVM, RF, XGB)}
\end{figure}

In Figure \ref{fig:3}, all models demonstrated a positive net benefit across a broad range of thresholds. Random Forest (Iso) consistently provided the highest net benefit at mid-range thresholds (0.3–0.6), while SVM (Iso) performed best at very low thresholds. In contrast, XGBoost (Iso) showed greater variability and a lower net benefit at higher thresholds.

\subsubsection{Fairness Analysis Across Demographic and Lifestyle Subgroups}
We calculated Accuracy, Precision, and Recall for different demographic and lifestyle groups to assess how well the model performs across various patient subpopulations. Table \ref{tab:subgroup_performance} below shows how well the model works for these specific groups.

\begin{table}[ht]
\centering
\caption{Model performance across demographic and lifestyle subgroups.}
\renewcommand{\arraystretch}{1.3}
\begin{tabular}{|c|c|c|c|c|}
\hline
\textbf{Group} & \textbf{Category} & \textbf{Accuracy} & \textbf{Precision} & \textbf{Recall} \\
\hline
Age & $<25$ & 0.692 & 0.571 & 0.800 \\
    & 25--35 & 0.909 & 0.867 & 0.765 \\
    & $>35$ & 0.897 & 0.800 & 0.667 \\
\hline
BMI & Normal & 0.909 & 0.733 & 0.846 \\
    & Overweight & 0.818 & 0.778 & 0.636 \\
    & Obese & 0.889 & 1.000 & 0.750 \\
\hline
Pregnancy & Non-pregnant & 0.892 & 0.765 & 0.765 \\
          & Pregnant & 0.853 & 0.800 & 0.727 \\
\hline
Marital Duration & $\leq$5 yrs & 0.846 & 0.667 & 0.667 \\
                 & $>$5 yrs & 0.897 & 0.833 & 0.789 \\
\hline
Exercise & No & 0.878 & 0.778 & 0.700 \\
         & Yes & 0.885 & 0.778 & 0.875 \\
\hline
Fast Food & No & 0.883 & 0.545 & 0.750 \\
          & Yes & 0.875 & 0.938 & 0.750 \\
\hline
\end{tabular}
\label{tab:subgroup_performance}
\end{table}

The model performed best in women aged 25–35, consistent with peak PCOS prevalence, but showed reduced accuracy in those under 25, likely due to underdiagnosis presentation. BMI group revealed perfect precision and moderate recall in obese individuals, while strong performance in normal BMI highlighted the model’s ability to capture lean PCOS phenotypes. Performance declined slightly during pregnancy, reflecting hormonal influences on diagnostic patterns. Longer marital duration was associated with better results, possibly due to reproductive history. Lifestyle analysis showed higher recall in regular exercisers. Fast-food consumers exhibited higher precision, suggesting more distinct clinical profiles.

\section{\textbf{Discussion}}
To enhance clinical trust and transparency in our PCOS prediction model, we conducted two key analyses:
\begin{itemize}
    \item Model explainability using SHAP to understand how features influence predictions.
    \item Fairness audit across demographic and lifestyle subgroups to assess equity in performance.
\end{itemize}

We selected the Random Forest (RF) model for both explainability and fairness analysis due to its strong overall performance and seamless integration with SHAP \cite{PonceBobadilla2024SHAP}. RF achieved competitive calibration and discrimination metrics under both isotonic and Platt scaling methods. Under isotonic calibration, RF attained a reliable classification and well-calibrated probabilities. Also, RF outperformed other models in the Brier score.

\subsection{Model Explainability with SHAP}
We utilized SHAP (SHapley Additive exPlanations), a state-of-the-art interpretability technique, to identify how various patient features influenced the model’s decisions.
Key findings from global SHAP analysis include:
\begin{itemize}
   
\item Ovarian Follicle Count (Left/Right): Higher counts were strongly associated with an increased risk of PCOS, consistent with the Rotterdam criteria, which identify polycystic ovarian morphology as a diagnostic hallmark \cite{Rotterdam2004PCOS}.
\item Weight Gain and Body Weight: Both higher body weight and recent weight gain positively influenced predictions, matching the established link between obesity, hormonal imbalance, and insulin resistance in PCOS\cite{Barber2019ObesityPCOS}.
\item Irregular Menstrual Cycles: Irregular cycles emerged as an important diagnostic feature, reflecting disrupted ovulation, a core symptom of PCOS \cite{Rotterdam2004PCOS}.
\item Skin Darkening (Acanthosis Nigricans): Often indicative of insulin resistance, this feature reduced the likelihood of a “non-PCOS” prediction, signalling strong importance in model decision-making \cite{Higgins2008Acanthosis}.

\item Hormonal Features (FSH levels): Even subtle hormonal shifts, within clinically “normal” ranges, contributed to refining predictions, highlighting the complexity of endocrine involvement in PCOS \cite{Saadia2020LHFSH}.
\end{itemize}




This alignment between model behavior and clinical makes a better understanding for both clinicians and patients that the model reflects real-world medical patterns rather than acting as a “black box.”

\subsection{Fairness Audit Across Patient Subgroups}
We assessed whether the model maintained consistent performance across diverse demographic and lifestyle-defined groups.

\textbf{Key subgroup insights:} Overall, the model is generally fair and reliable. 
Model findings across subgroups showed that performance was strongest in women in their peak reproductive years, while younger women had lower recall, suggesting that early PCOS is harder to detect. Accuracy was good across all body weight categories, with lean women demonstrating particularly strong recall, highlighting that PCOS is not confined to obesity. A slight accuracy drop was in pregnant women, likely due to hormonal shifts that can mimic PCOS symptoms. Marital duration, used as a proxy for reproductive history, showed no major differences. Women who exercised regularly with higher recall, possibly reflecting more stable health profiles. Frequent fast-food consumers exhibited better precision, which may reflect clearer metabolic patterns.

However, younger women and those with atypical PCOS presentations (lean, non-obese, pregnant) may face increased risk of misclassification. This highlights the importance of tailoring diagnostic models to diverse populations to minimize bias and prevent missed diagnoses.

\subsection{Clinical Integration and Actionable Implications}
\subsubsection{Implications for clinical and research use:}

The prototype holds promise as a supplementary tool for PCOS screening. By providing an initial risk assessment and flagging potential disparities, it could guide patients toward timely clinical consultation, potentially reducing diagnostic delays. 

From a research perspective, the tool provides a framework for embedding fairness and explainability into health risk models for other conditions. The "What-If" functionality offers a novel means to explore how lifestyle changes might impact individual risk, informing personalized medicine strategies. Future versions could incorporate real-time data from wearables and expand dataset diversity to improve representativeness.

The embedded Rotterdam criteria \cite{practitioners_polycystic_nodate} module operationalizes the 2003 consensus guidelines, allowing users to see whether their profile meets the $\geq$2 of 3 threshold (oligo/anovulation, hyperandrogenism, polycystic ovarian morphology) \cite{smet2018rotterdam}. The table of "Supportive / Contextual Indicators" adds further context with AMH levels, LH/FSH ratios, and waist circumference clinically relevant for lifestyle or follow-up decisions, even if not diagnostic. The interactive "What-If" feature empowers users to adjust modifiable factors such as weight and exercise, promoting proactive health engagement.

\subsection{EHR Integration}
Embedding the prototype into electronic health record (EHR) systems (like Epic, Cerner) via HL7 FHIR \cite{nazarov_how_2025} could enable prospective risk stratification using routinely collected data such as BMI, follicle count, and menstrual history. A clinician dashboard with SHAP-based visualizations would enhance interpretability by directly linking predictions to clinical indicators. 
\subsection{Addressing Fairness Gaps}
To address subgroup disparities, models should incorporate age-sensitive markers like AMH and longitudinal cycle tracking for younger women ($<$25 years) \cite{piltonen2024utility}, add insulin sensitivity indicators (HOMA-IR) to capture lean phenotypes better \cite{amisi2022markers}, and use pregnancy-adjusted feature sets to reduce misclassification among pregnant women.

\subsection{Policy and Systemic Recommendations}
Our findings may inform guideline development by professional bodies 
to emphasize early screening in younger and non-obese women. The model’s reliance on clinical and lifestyle features makes it scalable in low-resource settings, where costly hormonal assays are less feasible. Deployment must address privacy compliance 
clinician interpretability, and equitable access. Health economic assessments are warranted to determine cost-effectiveness, particularly in underserved populations.

\subsection{Limitations}
This study used a single dataset with limited demographic diversity, which may affect generalizability. Subgroup fairness audits were based on discretized variables ( BMI categories, age groups, etc.). Although RF was selected for interpretability, comparisons of multi-modal approaches (imaging, detailed hormonal assays, etc) were not performed. Our fairness and calibration analyses focused on group-level performance, not individual-level uncertainty or intersectional subgroup interactions.

The prototype depends on user-entered data, introducing potential input inaccuracies.

\subsection{Conclusion}
The "Fairness-Aware, Explainable PCOS Risk Tool" is a pioneering effort to merge accessibility, equity, and transparency in PCOS risk assessment. Although limited by the scope and challenges of the deployment of the dataset, its potential to empower users and inform research underscores its value as a foundation for equitable AI in women’s health.

 Generated by IEEEtran.bst, version: 1.14 (2015/08/26)

\end{document}